\documentclass[sn-mathphys,Numbered]{sn-jnl}

\usepackage{graphicx}%
\usepackage{multirow}%
\usepackage{amsmath,amssymb,amsfonts}%
\usepackage{amsthm}%
\usepackage{mathrsfs}%
\usepackage[title]{appendix}%
\usepackage{xcolor}%
\usepackage{textcomp}%
\usepackage{manyfoot}%
\usepackage{booktabs}%
\usepackage{algorithm}%
\usepackage{algorithmicx}%
\usepackage{algpseudocode}%
\usepackage{listings}%

\theoremstyle{thmstyleone}%
\theoremstyle{thmstyletwo}%

\theoremstyle{thmstylethree}%

\begin{document}
	\title[Assessing the Transformative Impact of Generative AI on Higher Education]{The Evolution of Learning: Assessing the Transformative Impact of Generative AI on Higher Education} 

\author*[1]{\fnm{Stefanie} \sur{Krause [\url{https://orcid.org/0000-0002-1271-7514}]}}\email{skrause@hs-harz.de}

\author[1]{\fnm{Bhumi Hitesh} \sur{Panchal}}
\equalcont{These authors contributed equally to this work.}

\author[1]{\fnm{Nikhil} \sur{Ubhe}}
\equalcont{These authors contributed equally to this work.}

\affil*[1]{\orgdiv{Department of Automation and Computer Science}, \orgname{Harz University of Applied Sciences}, \orgaddress{\street{Friedrichstrasse 57 - 59}, \city{Wernigerode}, \postcode{38855},  \country{Germany}}} 


\abstract{Generative Artificial Intelligence (GAI) models such as ChatGPT have experienced a surge in popularity, attracting 100 million active users in 2 months and generating an estimated 10 million daily queries. Despite this remarkable adoption, there remains a limited understanding to which extent this innovative technology influences higher education. This research paper investigates the impact of GAI on university students and Higher Education Institutions (HEIs).
	The study adopts a mixed-methods approach, combining a comprehensive survey with scenario analysis to explore potential benefits, drawbacks, and transformative changes the new technology brings.
	Using an online survey with 130 participants we assessed students' perspectives and attitudes concerning present ChatGPT usage in academics. 
	Results show that students use the current technology for tasks like assignment writing and exam preparation and believe it to be a effective help in achieving academic goals. 
	The scenario analysis afterwards projected potential future scenarios, providing valuable insights into the possibilities and challenges associated with incorporating GAI into higher education. 
	The main motivation is to gain a tangible and precise understanding of the potential consequences for HEIs and to provide guidance responding to the evolving learning environment. 
	The findings indicate that irresponsible and excessive use of the technology could result in significant challenges. Hence, HEIs must develop stringent policies, reevaluate learning objectives, upskill their lecturers, adjust the curriculum and reconsider examination approaches.
}


\maketitle

\section{Introduction}\label{sec1} 
New technologies have continuously revolutionized conventional teaching and learning approaches, causing fundamental changes in the educational system \cite{garcia2023perception}. Large Language Models (LLMs) have been utilized to create powerful chatbots, enabling them not only to comprehend and interpret human language input but also to generate content \cite{wei2023leveraging}. 
With the lastest developments in the field of transformer models, which are the basis for LLMs, the capabilities  of LLMs improved tremendously \sloppy \cite{transformer2022rapamycin}. 
Generative Pre-Trained Transformer (GPT) is such a model that is capable of generating human-like response text \cite{openai2023gpt4}. 
ChatGPT\footnote{\url{https://chat.openai.com/}} launched in November 2022 by OpenAI, can handle a variety of text-based requests, including question answering with near human level performance \cite{krause2024commonsense} and other difficult tasks like even passing one of the most difficult exams of Warton Business School with a B to B- grade \cite{terwiesch2023would}. 
These example illustrate the significant capabilities of the new GAI tool. However, there are other GAI tools the main aspect of ChatGPT's success lies in its public availability, user-friendliness and remarkable versatility across a wide range of applications and tasks. 
Despite the ongoing evolution of the foundational language model, it has advanced to GPT-4 Turbo in the paid version, we consider the free version, ChatGPT, for our specific focus.

In response to the media attention ChatGPT garnered, HEIs exhibited diverse reactions to these developments. Among the top 500 universities in the Quacquarelli Symonds (QS) World University Rankings published in 2022, the top 43 universities have adopted a policy deciding to ban ChatGPT by prohibiting the use of ChatGPT or any other AI tools in exams unless specifically authorized \cite{xiao2023waiting}. 
This will only serve to perpetuate and exacerbate the equity gaps that already exist \cite{bozkurt2023challenging,warschauer2023affordances}.
In contrast, we should adjust education to the new technology rather than banning it \cite{tlili2023if,meyer2023chatgpt}. 
Both teachers and students need to learn how to apply GAI responsibly \cite{sharma2022chat}.

The use of generative AI in education involves ethical issues, such as the possibility of students using it in an illicit or dishonest manner \cite{qadir2023engineering}, or how the technology may be used for inappropriate monitoring, control and evaluation \cite{jahn2019denkimpuls}. Furthermore, ChatGPT supports conclusions with factual arguments but may make mistakes by overemphasizing events and overlooking meta text, resulting in partial interpretations of news reports or statements \cite{kocon2023chatgpt}.
As the GPT systems are mostly trained for languages with abundant resources, such as English, it delivers highly accurate and competitive translation outputs but these models, still struggle with underrepresented languages \cite{hendy2023good}. 
For students as well as lecturers arise different opportunities and threats with GAI tools. 
The extent to which this emerging technology will transform education remains uncertain. 

There are a few bachelor or master programs in the field of Information Technology where lecturers are teaching AI as part of their curriculum, e.g., in the bachelor program \textit{AI Engineering}\footnote{\url{https://www.ai-engineer.de/}} \cite{krause2023entwicklung}. However, students of other disciplines need to profit from GAI as well. 
Therefore, we aim to discuss the educational implications of GAI models like ChatGPT in this research paper. 

The remainder of the paper is structured as follows: Section 2 highlights the related work, followed by Section 3 the description of our research questions. In Section 4, we present the methodology and our survey and scenario analysis. Then, in Section 5, we discuss the  main results and future research, and in Section 6, we draw together the key conclusions.\\
\\
\noindent
The \textbf{main contributions} of our paper are: 
\begin{enumerate}
\item Understanding how students use the hyped GAI tool ChatGPT
\item Exploring potential positive and negative implications for students and HEIs
\end{enumerate}

\section{Related Work} 
In recent years, numerous large language models like BERT \cite{Devlin2018}, RoBERTa \cite{Liu2019}, and LaMDA \cite{thoppilan2022lamda} and BART \cite{lewis2019bart} have been developed. 
However, only with the widespread popularity of ChatGPT this new technology significantly influenced and transformed our daily lives. 
Given the substantial impact of this disruptive technology on the education sector, it is crucial to grasp the opportunities, risks, and essential changes that HEIs must face. 
Atlas \cite{atlas2023chatgpt} mentions ways in which ChatGPT can be used in higher education, e.g., for automated essay scoring, research assistance, language translation, helping professors in creating their syllabus, quizzes, and exams, generating reports, email and chatbot assistance, etc. 
Rahman and Watanobe \cite{rahman2023chatgpt} stated that ChatGPT is a versatile tool that serves as an excellent assistant for learners, supporting them in understanding complex programming problems and generating an average accuracy of 85.42\%. 
In a comparative study between ChatGPT and Google Search on search performance and user experience Xue et al. stated that ChatGPT improves work efficiency, excels in answering straightforward questions and offers a positive user experience, but it may hinder further exploration and might not always outperform traditional search engines \cite{xu2023chatgpt}.

In addition to the numerous educational opportunities associated with ChatGPT, there has also been research conducted on its drawbacks and challenges.
Hosseini et al. queried ChatGPT and suggested that any section written by an natural language processing (NLP) system like ChatGPT should be checked by a domain expert for accuracy, bias, relevance, and reasoning and if it contains errors or biases \cite{hosseini2023using}. Furthermore, coauthors need to be held accountable for its accuracy, cogency, integrity. 
Mhlanga analysed in his literature review a responsible and ethical usage of ChatGPT \cite{mhlanga2023open}. Stating, educators must inform students about data collection, security measures and AI's limitations to foster critical thinking. 
Liu and Stapleton observed in their exploratory study between two groups of students, that conventional prompts in high-stakes English tests led to better performance, but experimental prompts in behavioral economics fostered diverse language use and enhanced critical thinking, highlighting the potential trade-off between standardized testing and cognitive skill development \cite{liu2018connecting}.

There has been limited research into students' viewpoints regarding GAI tools such as ChatGPT and the implications they may have for HEIs. 
Perera and Lankathilaka suggest using ChatGPT as a supplementary tool and not a replacement for human researchers, ensuring it is used under the supervision of experts by reinstating proctored, in-person assessments to reducing reliance on ChatGPT \cite{perera2023ai}. 
Cotton et al. mention ways of detecting work written by ChatGPT, firstly by looking for language irregularities or inconsistency that can indicate chatbot authorship \cite{cotton2023chatting}. Secondly, by checking for proper citations and references, and thirdly a lack of originality, factually inaccurate content, and error-free grammar. 
In contrast to the previous theoretical approaches, our method begins by recognizing how university students truly feel about using GAI tools like ChatGPT in their education.  Additionally, we consider possible scenarios for HEIs, to analyse implications and provide recommendations.

\section{Research Questions}  
This study attempts to investigate the effects of GAI on university students' education.
The first research question (RQ1) examines potential changes to student behavior and performance as a result of irresponsible usage of GAI. 
The aim is to comprehend the advantages and disadvantages of the usage of the prominent GAI tool ChatGPT from the perspective of students first, before considering the impact on educators and HEIs (RQ2).
We seek to explore how GAI could potentially revolutionize conventional teaching approaches by examining how is current usage in the educational context. 
This study aims to contribute to the ongoing dialogue on the integration of AI in higher education and to inform educators and policymakers about the implications to the HEIs  learning environment. With this consideration, we formulated the following two research questions: \\
\\
\noindent
\textit{\textbf{RQ1.} What are potential benefits and drawbacks for students using GAI for educational purposes?} \\
\textit{\textbf{RQ2.} What potential consequences or effects does GAI have on HEIs?}
\section{Methodology}
\begin{wrapfigure}{r}{0.38\textwidth}
\vspace{-20pt}
	\includegraphics[width=0.37\textwidth]{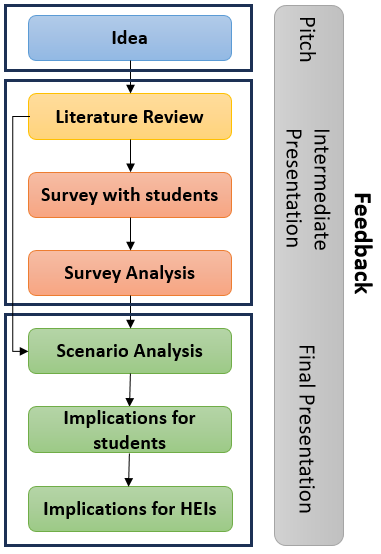}
	\caption{Our research process.}
	\vspace{-15pt}
	\label{figmethodology}
\end{wrapfigure}
To achieve a deep grasp of the subject, a mixed-method approach is used.  In our research, the quantitative method is a survey and the qualitative part a scenario analysis. Our complete methodology is presented in Figure \ref{figmethodology}. 
The validation process conducted during our research project incorporated a total of three feedback loops to ensure a high quality of our ideas, method and results.
Our feedback group consisted of around 20 students and two university professors, which were the audience of your pitch, intermediate and final presentations. 
Initially, the idea, motivation and proposed method was introduced through a pitch, setting a well-defined objective for the research. 
The second step involved a literature review, as well as carefully considering the optimal formulation of the survey questions to elicit the most valuable insights from students. The imperative was to pose queries encompassing both positive and negative aspects, maintaining impartiality throughout the process to capture a holistic understanding of the impact of GAI on academia. 
Data was collected through structured questionnaires via Google Forms\footnote{\url{https://www.google.com/forms/about/}}, ensuring participants anonymity and confidentiality. The results of the survey analysis were then presented in the intermediate presentation seeking feedback from the same audience as the pitch presentation. 

Afterwards we conducted a scenario analysis using the findings for our survey as well as insights from the literature review to investigate possible future paths for GAI integration in education. 
The development of credible scenarios considered the two main uncertainty dimensions the \textbf{frequency} and \textbf{responsibility} of students in the usage of GAI.
This refers to the ethical and conscientious utilization as students use GAI responsibly they verify critical information from reputable sources and ensuring accountability as well as academic integrity. 
The best case, worst case and base case scenario has been developed using these two uncertainty dimensions. 
These different scenarios offer a glimpse of possible outcomes and repercussions, assisting in the comprehension of the opportunities and difficulties presented by the integration of GAI technology in higher education settings.
This methodology provides comprehensive knowledge on the effect of GAI on university students' education by integrating survey data analysis with scenario prediction. It facilitates the exploration of diverse perspectives and the identification of potential trends, which helps us to provide recommendations for teachers.

\subsection{Survey Design}
The questionnaire was segmented into three sections. The initial section encompassed general information regarding the students' backgrounds, the second section inquired about the advantages and disadvantages of ChatGPT use, while the final section sought the students' expectations from their lecturers or HEIs. 
The survey was conducted using Google Forms due to its self-analysis features and user-friendly interface, which made result interpretation more straightforward. Subsequently, the responses were exported to an Excel spreadsheet for evaluation and data visualization. The survey link was distributed to students through the university's weekly newsletters and through networking among colleagues and different platforms.

\subsection{Analysis of the Survey Results}
This study revolved around student's interaction with the GAI tool ChatGPT.
Hence, the fundamental question was wheter the participants have actually used ChatGPT to provided the basis for comprehending and interpreting their responses effectively. Later on we evaluated the usage frequency more in detail. 
Out of the total pool of 130 participants, the survey focused its attention solely on the subset of 115 participants who affirmed their usage of ChatGPT.  
This selection allows a consistent and coherent examination of data throughout the subsequent stages. In contrast, the remaining 15 participants indicated a lack of engagement with ChatGPT, leading to the exclusion of their participation in this specific study. 

\begin{figure}[h!]
\centering
\includegraphics[scale=0.8]{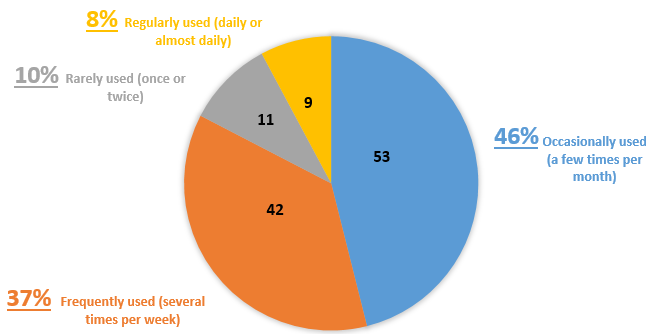}
\caption{Frequency how often students use ChatGPT.}
\label{figfrequency}
\end{figure}

The survey also sought to evaluate the frequency with which participants utilize ChatGPT for their study routines. Notably, the findings showcased a diverse pattern of usage. Approximately 46\% of participants, reported using ChatGPT a few times per month, indicating a periodic reliance on the tool. Additionally, 37\% of participants revealed a more frequent pattern of usage, relying on ChatGPT several times per week (see Figure \ref{figfrequency}). These results underscore the versatility of ChatGPT, catering to a variety of study habits and preferences among the participants.

Participants were asked to select various tasks for which they employed ChatGPT's assistance. The majority opted for the following tasks:
\begin{enumerate} 
\item Basic research or fact-checking
\item Generating ideas or brainstorming
\item Essay or assignment writing
\item Exam Preparation
\item Studying specific topics or concepts
\end{enumerate}	
\begin{figure}[h]
\centering
\vspace{-10pt}
\includegraphics[scale=0.58]{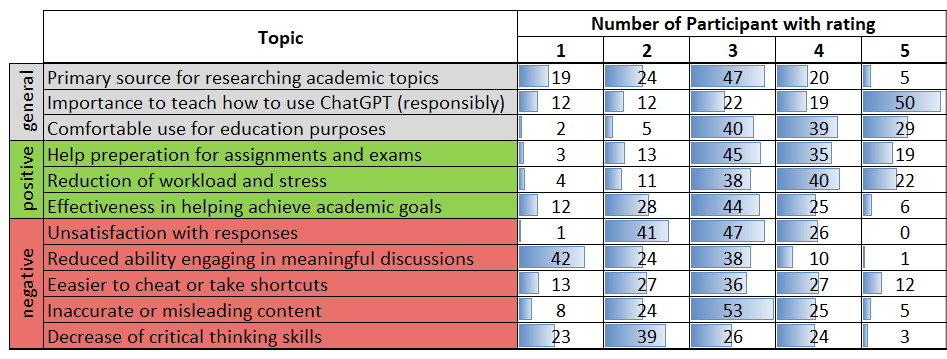}
\caption{Survey results clustered in general topics, positive and negative aspects. We used a 5 likert scale with 5 representing the highest possible approval. Total number of participants rating, mean and standard derivation (sd) are presented.}
\label{figfulltable}
\end{figure}
All survey results using a 5 likert scale are summarized in figure \ref{figfulltable}. 
The student survey regarding their experiences with ChatGPT revealed several key insights. The majority of participants rated ChatGPT's user comfort high (94\%) and found it more helpful than other resources (63\%), with many considering it their primary source (62\%). The majority of students use ChatGPT a few times per week or month. 
While 63\% were unsatisfied with the answers ChatGPT provided, 87\% believed it could reduce their workload and 65\% thought it could aid in achieving academic goals. Interestingly, more than 50\% did not feel that ChatGPT reduced their engagement, and 60\% did not perceive a negative impact on problem-solving skills. However, 65\% admit the tool  made cheating easier. The content generated also raised concerns, with 72\% rating it as misleading. Nevertheless, 57\% were open to using ChatGPT for exams or assignments. Regarding responsible use, 44\% rated instructors' efforts highly. Despite some reservations, a majority of 84\% of participants would recommend ChatGPT, primarily for brainstorming purposes, while stressing the need for fact-checking and corroborating information from other sources.

\subsection{Scenario Analysis}
Scenario analysis is a technique used to assess the potential outcomes of different situations or events, allowing individuals or organizations to make informed decisions based on a range of possibilities \cite{kosow2008methods}.
It involves creating various scenarios that represent different future states or conditions, each with its own set of assumptions and implications \cite{de2000brief}. The goal is to understand the potential risks, opportunities, and impacts associated with each scenario and provide recommendations how HEIs can prepare for the changing environment. 

There are various types of scenario analysis like single-point scenarios, best-case, worst-case, base-case scenarios, trend analysis, exploratory scenarios, black swan scenarios, and red-flag Scenarios \cite{kosow2008methods}.
For our research, we chose the best-case, worst-case, and base-case scenario to gain a well-rounded understanding of the potential outcomes, making us better equipped to improve our decision-making processes. 
The scenarios were the outcome of a systematic and iterative approach that included literature-based insights, survey analysis via students participation, and rigorous evaluation. This strategy made sure that the scenarios cover a wide range of probable outcomes and promote a thorough grasp of the likely future dynamics in the field of AI-enhanced education.

\begin{figure}[h]
\centering
\includegraphics[scale=0.485]{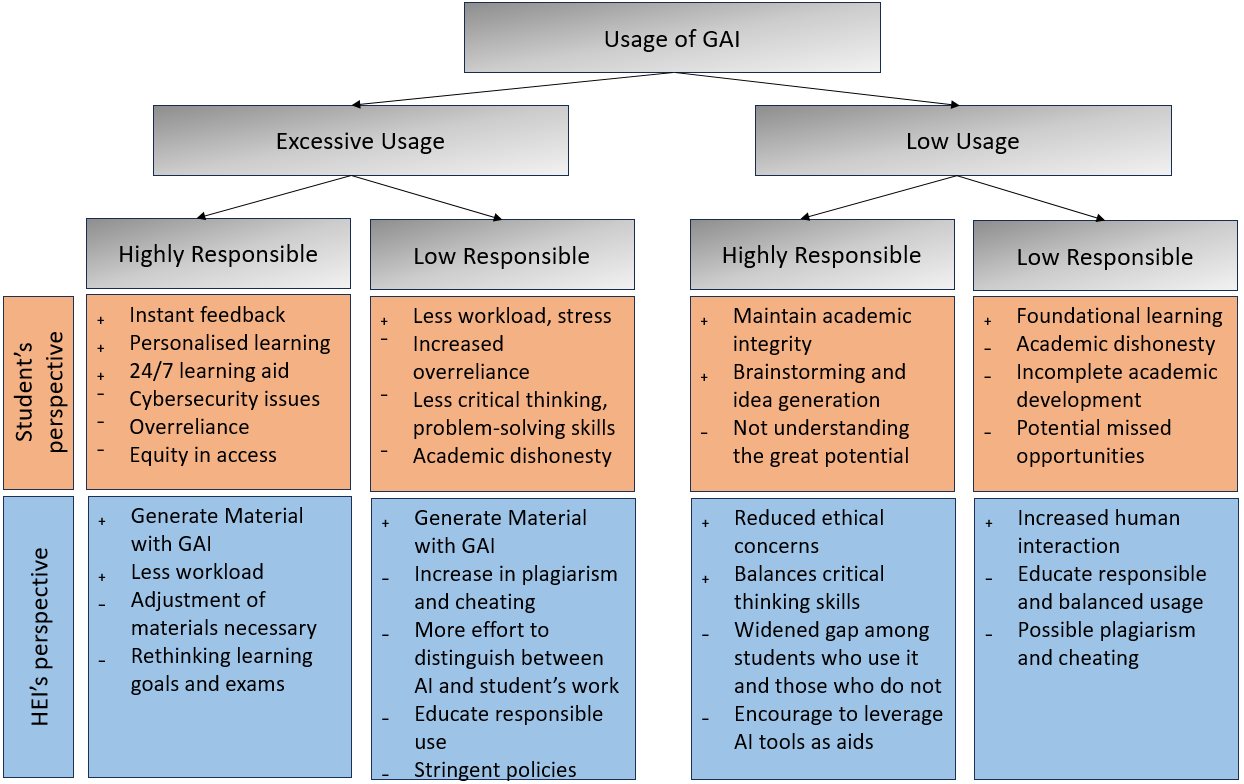}
\caption{Tree diagram showing advantages (+) and disadvantages (-) of the usage of GAI tools like ChatGPT based on frequency and level of responsibility for students as well as the perspective of HEIs. Four different scenarios result from the two main uncertenty dimensions: usage frequency and responsibility.}
\label{fig1}
\vspace{-1pt}
\end{figure}	
The tree diagram in Figure \ref{fig1} shows the different implications of GAI on students and HEIs depending on the two key uncertenties: the frequency of usage and level of responsibility of students. 
The tree ends at different future conditions depending on the path and thereby serves as a comprehensive visual representation for the four different scenarios that we found. 
We utilized four different categories that examine the effects of excessive and low usage together with responsible and irresponsible behavior. 
We characterize \textbf{low usage} as students employing GAI tools rarely, typically once or twice a month, whereas \textbf{excessive usage} implies daily or near-daily utilization of a GAI tool. \\
\\
\noindent
We characterize \textbf{highly responsible} behavior of students as:
\begin{itemize}
\item Awareness of limitations
\item Trusting human teachers over GAI
\item Maintaining academic integrity, ethical behavior
\end{itemize}
Conversely, we define \textbf{low responsbile or irresponsible} as:
\begin{itemize}
\item Lack of awareness of limitations 
\item Substituting human lecturers, overreliance
\item Academic dishonesty, unethical practices such as plagiarism and cheating
\end{itemize}

\begin{figure}[h!]
\centering
\includegraphics[scale=0.6]{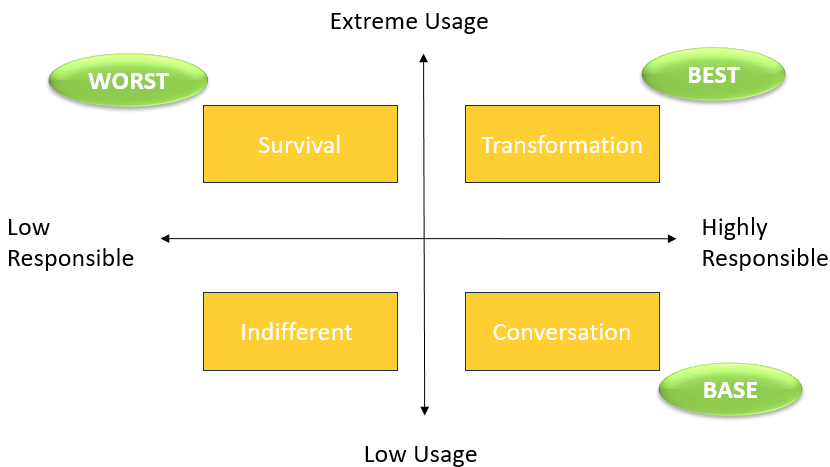}
\caption{Overview of four possible scenarios that are dependent on the amount and responsibility of usage of GAI tools. The bast-case, worst-case and base scenario are highlighted in green.}
\label{overview}
\end{figure}

After our literature review and survey analysis it becomes clear that the confluence of usage frequency and responsible behavior have significant effects on the possible future scenarios. However, there could be more factors (like government regulations, development of GAI tools) that influence future scenarios but these political and technological developments are difficult to predict. We rather want to focus on a micro level \cite{van2003updated} and consider the most important factors. 
Therefore, below we describe four scenarios, each presents another perspective on how students might utilize this technology and we explore the corresponding adjustments that teachers and professors may need to make in order to effectively integrate GAI into their instructional strategies. An overview of our scenarios is presented in figure \ref{overview}.
\\
\noindent
\\
\textbf{Scenario 1: Transformation} \\
In this scenario, students use GAI extensively but very responsibly, leading to a range of positive outcomes.  
Students will have access to personalized learning experiences, with AI tutors capable of adapting to individual learning styles, providing instant feedback, and generating tailored educational materials \cite{sharma2022chat}. 
It also serves as a powerful language learning aid, offering translation services and grammar/vocabulary explanations \cite{loos2023using}. 
With its 24/7 availability, students have access to assistance whenever they need it, even outside of regular school hours \cite{islam2023opportunities}. 
This revolution in education increases accessibility, especially for remote or underserved communities, as internet connectivity becomes more ubiquitous. 
GAI, alongside traditional teaching, would play a pivotal role in the transforming education \cite{gill2024transformative}. 
However, this scenario also raises concerns about the potential for over-reliance on technology \cite{sok2023chatgpt,fui2023generative}, issues of equity in access \cite{bozkurt2023challenging}, and the need for robust cybersecurity measures  \cite{wu2023unveiling} to protect sensitive student data. 
Additionally, the role of human educators shifts towards being facilitators and mentors, focusing on higher-order thinking skills \cite{iskender2023holy}, social and emotional learning, and the integration of technology into the curriculum. 
Lecturers can generate new material \cite{atlas2023chatgpt} or use GAI for learning assessment \cite{cotton2023chatting,gilson2023does} or to provide immediate feedback \cite{moore2022assessing} and thereby reduce workload \cite{sok2023chatgpt}, however adjustments of their material and exams becomes necessary. 
Rethinking of learning goals and how to measure these in an exam is of great importance, especially since our survey revealed that over half of the students want to be able to use ChatGPT even in exams. 
Upskilling competencies can become necessary for lecturers \cite{tlili2023if}. 
We described the best-case scenario representing a future where technology fundamentally reshapes the educational landscape, offering immense potential for accessible, personalized, and globally connected learning experiences. However, for this scenario HEIs need to make adjustments by, e.g., focusing on higher-order thinking skills, social and emotional learning, the integration of technology into the curriculum, upskilling lecturers, adjusting examation strategies.\\
\\
\textbf{Scenario 2: Conversation}\\
In this second scenario, students judiciously incorporate GAI into their educational experiences, striking a harmonious balance between AI-assisted learning and traditional pedagogical methods \cite{opara2023chatgpt}. 
Students use the new technology up to a few times per week, which is currently most realistic, as the usage frequency in our survey show that students use GAI tool like ChatGPT mostly occationally (a few times per week) 46\% and frequently (several times per week) with 37\%. 
GAI serves as a supplementary tool, offering valuable support for tasks such as concept clarification, and idea generation. 
Students exercise a high degree of responsibility, ensuring that the technology is utilized ethically and in adherence to academic integrity standards \cite{cotton2023chatting}. They engage with GAI in a manner that complements their existing learning strategies, leveraging its capabilities to enhance productivity and understanding. 
Human educators remain pivotal in education, offering guidance, mentorship, and critical thinking opportunities. They could utilize GAI to augment lessons, providing tailored explanations, creating practice exercises, and implementing personalized learning strategies \cite{mhlanga2023digital}. This collaborative approach enhances the overall educational experience. This scenario emphasizes the importance of responsible technology integration and encouraging students to leverage AI tools as aids rather than replacements for human-driven education. Lecturers should however encourage students to get to know the advantages of the AI tool, that there arise no gap between students using the technology and students who do not take advantage of the possibilities \cite{bozkurt2023challenging}. This is the base-case scenario, underscoring the value of a balanced approach, where both AI and human educators work in tandem to foster holistic and enriched learning experiences. 
We believe this scenario is the most realistic considering the survey result that 83\% of the participating students currently use GI tools like ChatGPT a few times a month or week. Since students use GAI responsibly this scenario is assumed to be the base case scenario.
In this scenario HEIs should integrate GAI tools into their lecturers to provide students with exposure to AI technologies. They could implement feedback mechanisms for students to provide input on their experiences with GAI tools, allowing for continuous improvement. They need to ensure that AI-powered resources and materials are accessible and promote digital literacy among students. Furthermore, HEIs regularly need to evaluate the impact and effectiveness of generative AI tools in achieving educational goals and be prepared to adapt and evolve strategies as AI technologies continue to develop and the usage behavior of students could change.\\
\\
\textbf{Scenario 3: Survival}\\
In this scenario, students depend extensively on GAI without exercising appropriate due diligence or responsibility in its utilization. The accessibility and ease of interaction with the AI system leads to over-reliance on its capabilities for various academic tasks \cite{sok2023chatgpt,fui2023generative}. Instead of actively engaging with course materials or seeking guidance from lecturers, students primarily rely on GAI as their main source for assignments, research papers, and similar tasks. Additionally, critical thinking and problem-solving skills may erode over time, as students become accustomed to instant answers and automated assistance reducing workload and stress \cite{limna2023use}.  Plagiarism becomes a prevalent issue, as students may submit content generated by GAI tools without proper attribution or original thought \cite{karthikeyan2023literature}. This scenario raises concerns about the erosion of academic integrity, with educational institutions grappling to detect and address instances of irresponsible GAI use \cite{yu2023reflection}. Furthermore, it highlights the potential for missed learning opportunities and diminished engagement in collaborative, interactive learning environments \cite{tlili2023if}. Educators and institutions are prompted to implement stringent policies \cite{fui2023generative}, educational campaigns, and technological safeguards to mitigate the negative consequences of this unbridled dependence on GAI. The responsibile of GAI needs to be incorporated in the curriculum to guide students how to use the technology in an appropriate manner. Lecturers need to teach limitations of GAI and punish academic dishonesty and unethical practices. This scenario presents the worst-case scenario serving a cautionary tale, emphasizing the need for balanced, responsible use of AI tools in education to preserve the integrity and effectiveness of the learning experience.\\
\\
\textbf{Scenario 4: Indifferent}\\
In this scenario, despite the availability of GAI as a valuable educational resource, students opt to use it sparingly and when in use then in an inappropriately manner. 
Rather than leveraging the tool for its intended purpose of assisting in the learning process, it is predominantly utilized for shortcuts, such as generating plagiarism or seeking immediate answers without genuine comprehension \cite{lo2023impact}. This minimal engagement with GAI leads to missed opportunities. It becomes necessary for the educators to teach how to use the technology appropriate. Furthermore, irresponsible use may lead to issues of academic integrity and dishonesty, as students may resort to unethical practices in their educational pursuits \cite{karthikeyan2023literature}. This scenario thus highlights the importance of promoting responsible and meaningful engagement with GAI in the curriculum, emphasizing its true value as an educational tool rather than abandoning it. HEIs need to teach limitations of GAI and punish academic dishonesty and on the other hand promote the advantages of responsible by showcasing positive examples in class. 
However, we will not focus on this scenario in more detail as it is not one of the three best-case, worst-case or base scenarios.
\subsection{Results of the Scenario Analysis}
As we investigate all scenarios, it becomes evident that there are some underlying positive and negative implications.
Academic dishonesty and overreliance of students and for lecturers more effort to distinguish between AI and students work but easy generation of new materials, are examples of elements that cut across multiple scenarios, influencing academic outcomes and interactions. 
The major positive implications include tailored educational teaching, instant feedback capability and 24/7 assistance resulting in an improved learning journey for the students. On the other hand, irresponsible use of technology would challenge academic integrity, leading to unethical practices like plagiarism and cheating. For HEIs, educating the right practices becomes vital going forward on this path of awareness to a responsible usage of technology among students. It is equally important to stress on the fact that regular physical interaction between the students and educators are necessary to maintain social and emotional learnings. Incorporating GAI tools into the traditional classroom concept is highly recommended as it brings many advantages. However, the behavior of students should be monitored as it might change over time. At the moment students use GAI rather moderate, and we believe to be currently in the scenario \textit{Conversation}, however this might change over time to a more extreme usage. In that case HEIs have to prepare for the scenarios \textit{Transformation} or \textit{Survival} depending on how responsibly students use GAI tools. 

Over all szenarios there occure important common topics. Independently from the specific scenarios we \textbf{recommend teacher in higher educational instutions} to:
\begin{itemize}
	\item integrate GAI in classes, especially teaching students a responsible GAI usage
	\item inform about limitations of GAI
	\item rethink learning goals and materials
	\item renew exams in cases where GAI usage is possible and decide whether it is permitted and if not how to make sure there is no cheating
\end{itemize}

\section{Discussion and Future Research} 
The examination of different levels of GAI utilization in conjunction with responsible or irresponsible conduct presents an comprehensive first exploration. The impact of GAI on education might take a longer time frame to fully unfold, and longitudinal studies could shed more light on the sustained effects of incorporating this technology in education. 
This line of inquiry could uncover the impact of excessive AI reliance, the consequences of misinformation and bias, psychological dynamics of human-AI interaction, as well as the development of ethical guidelines. 
Such discussions and future research hold the potential to guide responsible AI integration, inform policymaking, and ensure a balanced, empowered, and ethically sound coexistence with AI technologies.

While this study highlights the benefits perceived by students, such as improved assignment assistance, enhanced learning experiences and stress reduction we need to acknowledge certain limitations of our work. The sample size and demographics of survey participants were restricted, making it challenging to generalize the findings.  
Additionally, the study focuses solely on students' perspectives, neglecting educators' views in the survey. 

In conclusion, the research paper offers valuable insights but requires further investigation. By addressing the limitations through larger and more inclusive studies, involving educators' perspectives, and examining the long-term effects, researchers can gain a more comprehensive understanding of how to effectively leverage GAI to enhance education for students.

\section{Conclusion} 
In conclusion, this research paper explored the impact of GAI on students and HEIs through a comprehensive survey and scenario analysis. The findings shed light on the diverse experiences and perspectives of students using GAI as an educational tool. The interplay between usage frequency and responsible behavior greatly influences the outcomes of integrating GAI into education. Employing GAI with mindfulness can amplify its advantages while mitigating potential pitfalls. Overall, the research underscores the importance of a balanced approach when utilizing GAI. While it has the potential to enhance learning experiences and reduce workload, caution must be exercised in addressing concerns related to content reliability, overreliance and dishonesty of students. Instructors play a crucial role in teaching responsible usage, with many students acknowledging the significance of such guidance. However, lecturers have to adjust their materials and curriculum and rethink learning goals as well as their exams. Furthermore, HEIs have to increase their efforts to detect academic dishonesty and plagiarism, provide stringent policies and reduce of ethical concerns especially when students use GAI excessively with low responsibility. 
As the technology of GAI continues to evolve, further research and adaptability in educational strategies are essential to maximize the benefits of GAI while mitigating potential drawbacks. By understanding and addressing these impacts, we can effectively harness the power of GAI to enrich the learning experiences of students. \\
\\
\noindent
\textbf{Funding.} The research reported in this paper has been supported by the German Federal Ministry of Education and Research (BMBF) in the Programme \emph{Künstliche Intelligenz in der Hochschulbildung} under grant no.~16DHBKI010. \\
\\
\noindent
\textbf{Acknowledgements.} We thank all students who voluntarily participated in the survey and the participants in our feedback sessions. Furthermore, we like to acknowledge the insightful comments and constructive suggestions from Frieder Stolzenburg.


\bibliography{bibliography}

\end{document}